\documentclass[runningheads]{llncs}

\usepackage{eccv}

\usepackage{eccvabbrv}
\usepackage{graphicx}
\usepackage{booktabs}
\usepackage{amsmath}
\usepackage{mathabx}

\usepackage[accsupp]{axessibility}  

\usepackage[pagebackref,breaklinks,colorlinks,citecolor=eccvblue]{hyperref}

\usepackage{orcidlink}

\newcommand{\method}{Garment Attribute Manipulation with Multi-level Attention}
\newcommand{\short}{GAMMA}

\usepackage{array}
\newcolumntype{H}{>{\setbox0=\hbox\bgroup}c<{\egroup}@{}}

\begin{document}

\title{Garment Attribute Manipulation with Multi-level Attention} 

\titlerunning{Garment Attribute Manipulation with Multi-level Attention}

\author{Vittorio Casula \inst{1}\orcidlink{} \and
Lorenzo Berlincioni \inst{1}\orcidlink{0000-0001-6131-1505} \and
Luca Cultrera \inst{1}\orcidlink{0009-0003-2483-9927} \and
Federico Becattini \inst{2}\orcidlink{0000-0003-2537-2700} \and
Chiara Pero\inst{3}\orcidlink{0000-0002-5517-2198} \and
Carmen Bisogni\inst{3}\orcidlink{0000-0003-1358-006X} \and
Marco Bertini\inst{1}\orcidlink{0000-0002-1364-218X} \and
Alberto Del Bimbo\inst{1}\orcidlink{ 0000-0002-1052-8322}}

\authorrunning{V.Casula et al.}

\institute{University of Florence, ITALY, \email{name.surname@unifi.it} \and
University of Siena, ITALY, \email{name.surname@unisi.it} \and University of Salerno, ITALY, \email{nsurname@unisa.it}}

\maketitle

\begin{abstract}
  In the rapidly evolving field of online fashion shopping, the need for more personalized and interactive image retrieval systems has become paramount. Existing methods often struggle with precisely manipulating specific garment attributes without inadvertently affecting others. To address this challenge, we propose GAMMA (Garment Attribute Manipulation with Multi-level Attention), a novel framework that integrates attribute-disentangled representations with a multi-stage attention-based architecture. GAMMA enables targeted manipulation of fashion image attributes, allowing users to refine their searches with high accuracy. By leveraging a dual-encoder Transformer and memory block, our model achieves state-of-the-art performance on popular datasets like Shopping100k and DeepFashion. 
  
\end{abstract}

\section{Introduction}
\label{sec:intro}
In the digital age, online fashion shopping is becoming increasingly popular, with consumers seeking more personalized and interactive shopping experiences such as virtual try-on~\cite{morelli2022dress,han2018viton} or garment design~\cite{baldrati2023multimodal,baldrati2024multimodal}. 
In this context, fashion image retrieval has gained particular interest in recent years~\cite{10.1145/3636552}. Fashion Image Retrieval (FIR) identifies comparable products in pictures datasets based on customer preferences. FIR methods have to handle typical problems as variation in viewpoint, deformation, occlusion, and the abstract concept of similarity in fashion. 
To address these challenges and enhance user satisfaction, the integration of interactive image search in fashion is emerging as a powerful approach. 
Interactive image search \cite{9432822} allows users to actively refine and adjust their search queries in real-time, providing immediate feedback on attributes like color, style, and fabric. This dynamic interaction helps to overcome the inherent complexities in fashion image retrieval by enabling more precise and tailored results, ultimately creating a more engaging and user-centric shopping experience. By combining the strengths of FIR with the adaptability of interactive image search, the fashion industry can significantly improve the accuracy and relevance of product recommendations.
However, interactive image search presents several challenges, particularly when it comes to adapting search results based on user feedback \cite{Wu_2021_CVPR}. For example, a user may want to change the color of a T-shirt in search results, but this often results in undesirable changes in other attributes, such as sleeve type. This problem stems from the fact that visual representations are semantically intertwined, which limits the ability to precisely control search results. To overcome this limitation, some methods in recent years have relied on the disentangled representations approach of the attributes that identify the object \cite{hou2021ADDE,scaramuzzino2023attribute}. However, even if disentangled representations are a significant step towards effective interactive search, attribute manipulation techniques still need to be better explored. In our work we pursue this line of research by proposing \short{} a \method{} method that, starting from a disentangled representation, actively manipulates the features to retrieve fashion items that better match the desired attributes. The key component of our approach is a multi-stage attention-based architecture, trained to compute the target representation.

Our main contributions can be summarized as follows:
\begin{itemize}
\item We propose an architecture to perform fashion image retrieval by manipulating a query image according to a desired attribute manipulation.  
Our approach combines existing attribute-disentangled representations with a new transformer-based manipulation strategy.
\item Thanks to an attribute-based attention and to the modularity of our approach, we can compose multiple submodules into complex architectures by combining self and cross-attention blocks.
\item We conducted experiments on two popular fashion item datasets, Shopping100k and DeepFashion, reporting state-of-the-art results.

\end{itemize}
\begin{figure}[t]
    \centering
    \resizebox{1.05\linewidth}{!}{
    \includegraphics[]{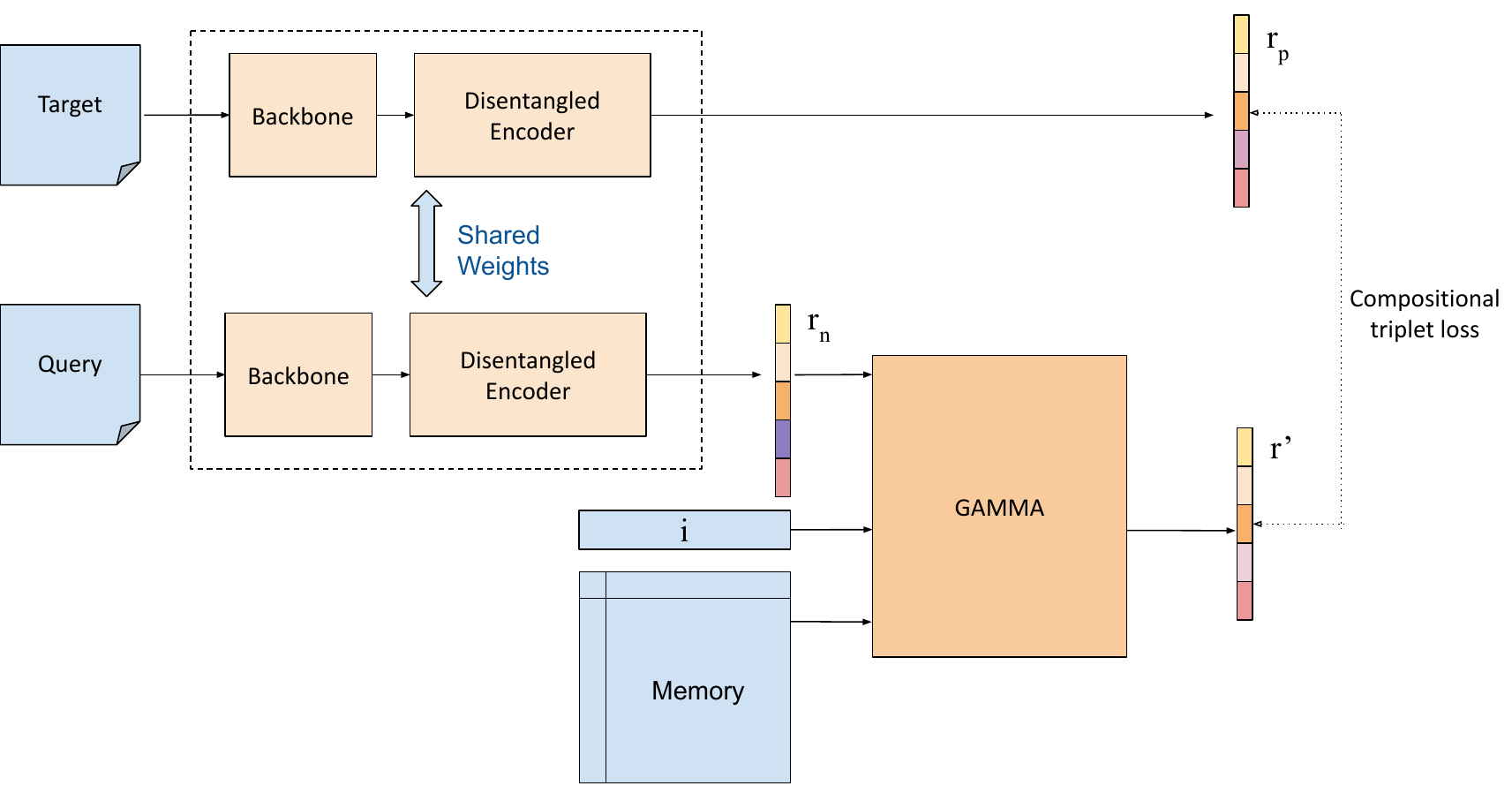}}
    \caption{Illustration of our model. The disentangled representation of the query image, $r_n$, is provided to GAMMA along with the manipulation indicator vector $i$ and the memory of prototype features $M$.}
    \label{fig:teaser}
\end{figure}

\section{Literature Review}
\label{sec:relworks}

\subsection{Fashion Image Retrieval}
The rapid expansion of e-commerce within the fashion industry has led to a significant demand for advanced solutions that enable customers to quickly identify and locate their desired fashion items. Image-based fashion retrieval systems have been developed to meet this need, allowing users to search for similar items using a reference image\cite{kovashka2012whittlesearch,park2019study}. This technological advancement has prompted a surge in research across various related fields, including fashion image retrieval~\cite{baldrati2024multimodal}, clothing detection~\cite{ge2019deepfashion2}, fashion recommendation~\cite{song2019gp,becattini2021plm}, fashion analysis~\cite{liu2021mmfashion}, and fashion synthesis~\cite{becattini2023transformer}. Current research focuses on areas such as cross-domain fashion retrieval, the detection of clothing, personalized fashion recommendations, and both fashion analysis and synthesis \cite{10.1145/3636552}. Han et al. \cite{han2017learning} introduced an end-to-end method that jointly learns visual-semantic embeddings and compatibility relationships among fashion items. They employed a bidirectional LSTM (Bi-LSTM) to sequentially predict the next item in an outfit, capturing the compatibility of items based on preceding ones. The advancement of attention mechanisms has enhanced models' ability to understand context, leading to their extensive use in recommendation systems \cite{chen2021crossvit, vaswani2021scaling}. Nevertheless, these studies largely concentrated on the broad similarity of clothing items, and did not address the requirements for more precise, fine-grained retrieval. To overcome this limitation, Ma et al. \cite{ma2020fine} proposed an Attribute-Specific Embedding Network (ASEN). This model concurrently learns multiple attribute-specific embeddings in an end-to-end framework, facilitating the accurate measurement of fine-grained similarities within the relevant attribute space. On the other hand, Jiao et al. \cite{jiao2022fine} developed a deep learning-based online clustering method to jointly learn fine-grained fashion representations for all attributes at both the instance and cluster levels. Recently, Xiao et al. \cite{xiao2024attribute} presented the Attribute-Guided Multi-Level Attention Network (AG-MAN). This method enhances feature extraction by capturing multi-level image embeddings and groups images with similar attributes into the same class, thereby addressing feature gaps.

\subsection{Attribute Manipulation}
With the expansion of online shopping, understanding how garments can be characterized by their components and/or attributes, which collectively describe the item, has become essential. Identifying and detailing these attributes has emerged as a significant research area. Abdulnabi et al. \cite{abdulnabi2015multi} proposed a joint multi-task learning algorithm using deep convolutional neural networks (CNNs) to predict attributes from images, facilitating the sharing of visual knowledge across different attribute categories. Similarly, Liu et al. \cite{deepfashion} introduced FashionNet, a deep learning model that extracts clothing features by concurrently predicting both attributes and landmarks. With the advent of Generative Adversarial Networks (GANs), the problem has been redefined as a task involving image generation \cite{zhu2016generative, ak2019attribute}. However, these methods frequently complicate the retrieval of real garments, as their efficacy is closely dependent on the quality of the generated images.
In response, Zhao et al. \cite{zhao2017AMNet} developed a memory-augmented Attribute Manipulation Network (AMNet) designed to modify image representations at the attribute level, rather than generating new images, thereby facilitating direct retrieval. Specifically, AMNet alters the intermediate representation of a query image to replace specified attributes with the desired ones.
The authors in \cite{ak2018FSN} employed memory-augmented networks to maintain prototypes of disentangled features. Building on this concept, they introduced FashionSearchNet (FSN), which utilizes a weakly supervised localization technique to extract attribute-specific regions. This approach filters out irrelevant features, thereby enhancing similarity learning. Additionally, FSN incorporates a procedure for region awareness, enabling it to effectively address queries related to specific regions. In \cite{de2023disentangling} instead the feature disentanglement is obtained using a contrastive learning approach. Similar variations using memory augmented networks are proposed in \cite{becattini2023fashion,de2021garment,de2021style}.
According to \cite{hou2021ADDE}, a common challenge in interactive retrieval is that user interactions can inadvertently affect other aspects. This issue arises because current methods generate visual representations that are semantically entangled within the embedding space, thus limiting control over the retrieved results. To address this problem, they proposed ADDE, which enables the networks to learn attribute-specific subspaces and achieve disentangled representations.
Bhattacharya et al. \cite{bhattacharya2022datrnet} introduced DAtRNet, an attribute representation network that focuses on attribute-level similarity to provide precise recommendations. This end-to-end framework effectively disentangles attribute features and processes them independently, thereby enhancing flexibility in managing single or multiple attributes.
In a similar vein, the authors in \cite{scaramuzzino2023attribute} adopted an attribute disentanglement technique that leverages attribute classifiers and gradient reversal layers. This method allows for the extraction of attribute-specific features while filtering out irrelevant characteristics.

\section{Methodology}
\label{sec:method}
The task of retrieving images based on a selected attribute of a query image can be formulated as the following function:
\begin{equation}
     f: (Q_{\text{img}}, q) \in \mathcal{I} \to \mathcal{S}_q(\mathcal{I})
\end{equation}

where \( Q_{\text{img}} \) is an image that can be viewed as a finite set of attributes \( \{a_1, a_2, \ldots, a_n\} \), and each attribute \( a_i \) can take on a finite number $L_i$ of values $v_i^{k=1,...,{L_i}}$. Additionally, \( q \) is an attribute manipulation indicator \((a_j, v^{-}_{j}, v^{+}_{j}) \), which is provided as query. Here $v_{j}^-$ indicates the value of attribute $a_j$ that should be removed and replaced with value $v_{j}^+$.
The function thus maps the query image and manipulation vector from the input image space $\mathcal{I}$ to its subset $\mathcal{S}_q(\mathcal{I})$ containing all images that have the same attributes as \( Q_{\text{img}} \) except for the attribute \( a_j \) given in the query, which must be equal to $v_{L_j}^+$, as defined in the manipulation $q$.


Formally, let 
\begin{equation}
     Q_{\text{img}} = \{(a_1, v_1^{k}), (a_2, v_2^{k}), \ldots, (a_n, v_n^{k})\} 
\end{equation}
where \( a_i \) are attributes and $v_i^{k'}$ are their corresponding values. Let  \( q = (a_j, v^{-}_{j}, v^{+}_{j}) \) be the manipulation defined in the query. Then,

\[
\mathcal{S}_q(\mathcal{I}) = \{ I \in \mathcal{I} \mid I = \{(a_1, v_1^{k'}), (a_2, v_2^{k'}), \ldots, (a_j, v^{+}_{j}), \ldots, (a_n, v_n^{k'})\} \}
\]

where \( v_i^{k'} = v_i^{k} \) for \( i \neq j \).

\subsection{Attribute Disentanglement}

Given the above definition, it is clear that the first step of the method involves the disentangled representation of the image as a set of attributes.

To reach this goal, we adopt the method introduced by Hou et al. in \cite{hou2021ADDE}. 
The ADDE model (Attribute-Driven Disentangled Encoder) has the aim of creating $n$ classifiers, where $n$ is the number of attributes $a$ of an Image $Q_{img}$. Firstly, a CNN architecture is used as a feature extractor, in this case AlexNet \cite{NIPS2012_c399862d} has been choose for the 
purpose. The initial image $Q_{img}$ is thus transformed in the image representation $R(Q_{img})$. At this point, $R(Q_{img})$ is fed to $n$ networks $f2_i$ composed of two fully connected layers each. Each of the networks is associated to one of the attributes of the image, for this reason, we obtain $(f2_1, f2_2,...,f2_n)$ networks that generate $n$ representations
\begin{equation}
    r_1=f2_1(R(Q_{img})), r_2=f2_2(R(Q_{img})), ..., r_n=f2_n(R(Q_{img}))
\end{equation}
The final predictions are made by a softmax fully connected layer obtaining $\overline{y}_1, \overline{y}_2, ..., \overline{y}_n$.
The loss used is a cross-entropy:
\begin{equation}
    L=-\sum_{k=1}^{N} \sum_{i=1}^{n} log(p(y_{k,i}, \overline{y}_{k,i})
\end{equation}
where $N$ is the number of samples in the training stage, $n$ is, as previously introduced, the number of attributes, and $y_{k,i}$ is the ground truth label of the $k-th$ image $Q_{img_k}$ for the $i-th$ attribute. 
The disentangled representation of the image $Q_{img}$ is finally obtained by concatenating the embeddings inferred for each attribute
$r=r_1,...,r_n$, with $r$ $\in$ $\mathbb{R}^{nL}$ with $L$ the dimension of each attribute-specific embedding. 
We leverage ADDE as a starting point for building our proposed approach, integrating our attribute manipulation architecture \short{} after the disentangled feature extractor of \cite{hou2021ADDE}, as shown in Figure \ref{fig:teaser}.

\begin{figure}[t]
    \centering
    \resizebox{1.05\linewidth}{!}{
    \includegraphics[]{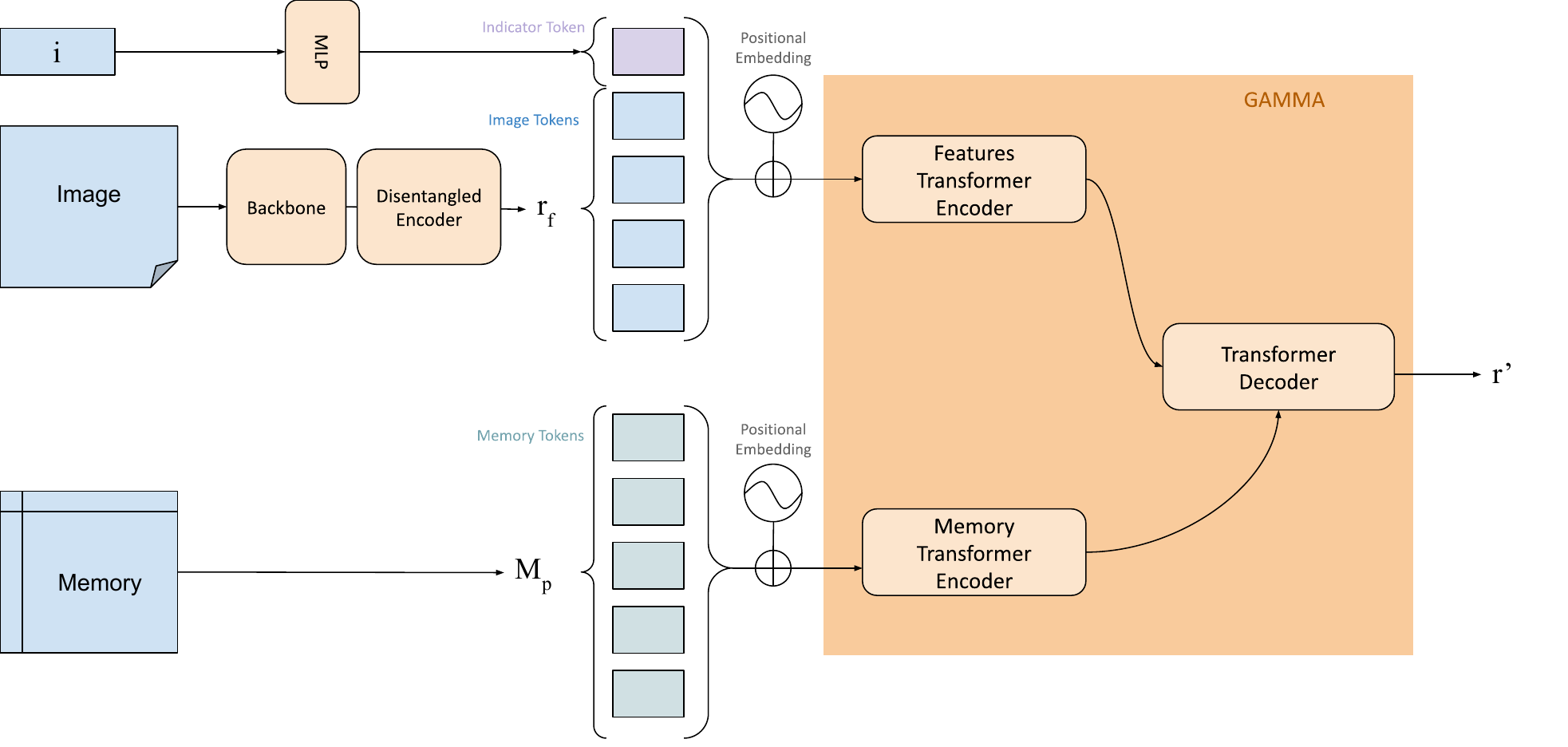}}
    \caption{GAMMA internal architecture illustration.}
    \label{fig:diagram}
\end{figure}

\subsection{\method{}}

We propose \short{}, an architecture for garment attribute manipulation that works with disentangled features and leverages a hierarchy of attention mechanisms to transform them to reflect a desired change.
The main idea behind \short{} is that we can represent a feature-disentangled fashion item as a set of attribute tokens. This allows us to feed them into transformer-like modules and exploit self and cross-attention schemes.

Our proposed architecture is depicted in Figure~\ref{fig:diagram} and works as follows. After the disentangled representation has been extracted from the backbone and the feature encoders, we feed them to a transformer encoder. Along with the features $r_1,...r_n$ we also inject as input a representation of the desired manipulation $q$.
The manipulation is first represented as an indicator vector \( \mathbf{i} \in \{-1,0,1\}^{L_{tot}}\). The vector contains all zeros, a +1 and a -1, where the negative entry indicates the attribute value we want to remove and the positive one the value we want to add. The indicator vector $\textbf{i}$ has as many element as the total number of attribute values, i.e. $L_{tot}=\sum_{i=1}^{n} L_i$, with $L_i$ specifying the number of values that the attribute $i$ can assume. By construction, both the +1 and the -1 are going to be set within the subvector relative to the same attribute $j$ that we want to modify.
In order to feed this information to the model, we project it into the token space with a multi-layer perceptron with two layers, ReLU activation and layer normalization~\cite{ba2016layer}.

At this point, we apply a sinusoidal positional encoding $PE(\cdot)$ to the set of tokens composed of the projected indicator vector and the attribute features and feed them to a transformer encoder:

\begin{equation}
    \mathbf{r_F} = \text{Encoder}(PE([\mathbf{r_1}, \mathbf{r_2}, ..., \mathbf{r_n}, MLP(\mathbf{i})]))
\end{equation}

The purpose of the transformer encoder is two-fold. First, it focuses the computation on the embedding corresponding to the attribute \( a_j \) from those of other attributes. This allows the model to remove the information relative to the negative value and start manipulating the feature towards the value to be added.
Second, the self-attention of the transformer encoder relates features to each other, which, although disentangled, might still carry traces of information belonging to other attributes.
The resulting transformed features $\mathbf{r_F}$ are passed to the next layer.

In parallel to the first encoder, another transformer encoder processes additional information, taking inspiration from prior work such as \cite{hou2021ADDE, scaramuzzino2023attribute, zhao2017AMNet}. In fact, we build a memory of prototype features. The memory block can be defined as a matrix $M \in \mathbb{R}^{nL \times L_{tot}}$.
Prototype embeddings for every attribute value are stored in $\mathbf{M}$. For instance, a prototype embedding for every distinct pattern in the dataset is given for the pattern property, similarly, a prototype will be present for each category or color. The memory block is initialized by averaging the disentangled embeddings of training data having the same attribute value. Formally, the memory block matrix has the following form:
\begin{equation}
    \mathbf{M}= \begin{pmatrix}
    p_1^1 & ... & p_1^{L_1}& 0 & ...& 0&0&...&...\\
    0 & ... & 0& p_2^1 & ...& p_2^{L_2}&0&...&0 \\
    ... & ... & ...& ... & ...& ...&...&...&... \\
    0 & ... & 0& 0 & ...& 0&p_n^1&...&p_n^{L_n}
    \end{pmatrix}
\end{equation}
where $p_i^j$ is the prototype embedding of the attribute $i$, having the $j-th$ attribute value. This is similar to the memory block of \cite{hou2021ADDE, scaramuzzino2023attribute}.
For each prototype, we extract only the non-zero values, i.e. ignoring the values not belonging to the feature of the corresponding attribute. This leaves us with a set of single-attribute prototypes $\mathbf{M_p} = [p_1^1, ..., p_1^{L_1}, p_2^1, ..., p_2^{L_2}, ..., p_n^1, p_n^2,...,p_n^{L_n}]$, which we feed to the second transformer encoder, after applying a positional encoding.
Similarly to the first transformer, this layer relates the features to each other and transforms them before passing them to the next layer.

\begin{equation}
\mathbf{r_M} = \text{Encoder}(PE(\mathbf{M_p}))
\end{equation}

Finally, the outputs of the two encoders are processed via cross-attention. We exploit a Transformer decoder to relate the transformed attribute features $\mathbf{r_F}$ with the transformed prototypes $\mathbf{r_M}$. Since $\mathbf{r_F}$ contains also the transformed indicator vector, encoding the desired transformation, \short{} can extract relevant information about the appearance of the desired attribute value from the prototypes and transform the input features accordingly.

The final manipulated representation is thus obtained as:
\begin{equation}
\mathbf{r'} = \text{Decoder}(\mathbf{r_F},\mathbf{r_M})
\end{equation}

\subsubsection{Training}
Following \cite{hou2021ADDE}, we employ two distinct loss functions to ensure effective feature manipulation. These loss functions are Compositional Triplet Loss and Label Triplet Loss, each addressing specific aspects of the model's training objectives.

\begin{itemize}

\item \textbf{Compositional Triplet Loss}:
The Compositional Triplet Loss function is designed to enforce the desired proximity between the manipulated embedding \( \mathbf{r'} \), the target prototype embedding \( \mathbf{r_p} \), and a negative sample embedding \( \mathbf{r_n} \). This loss ensures that the manipulated features are closely aligned with the target while being distinct from irrelevant or incorrect modifications. Formally, it is defined as:

\begin{equation}
    \mathcal{L}_{comp} = \max(0, d(\mathbf{r'}, \mathbf{r_p}) - d(\mathbf{r'}, \mathbf{r_n}) + \alpha)
\end{equation}

where \( d (\cdot)\) is the $L_2$ distance metric, \( \alpha \) is a margin that separates positive and negative samples, \( \mathbf{r'} \) is the transformed embedding, \( \mathbf{r_p} \) is the target embedding, and \( \mathbf{r_n} \) is a negative sample. This loss encourages the model to bring the manipulated embedding closer to the desired target while distancing it from embeddings that do not align with the intended modification.




\vspace{0.5cm}
\item \textbf{Label Triplet Loss}: The Label Triplet Loss function is designed to ensure that the manipulated embedding \( \mathbf{r'} \) aligns with the correct target attributes while diverging from incorrect or negative attributes. This loss helps in fine-tuning the attribute-specific manipulations to adhere to the desired labels. It is given by:

\begin{equation}
    \mathcal{L}_{label} = \max(0, d(\mathbf{r'}, \mathbf{v_p}) - d(\mathbf{r'}, \mathbf{v_n}) + \beta)
\end{equation}

where \( \mathbf{v_p} \) is the positive attribute value corresponding to the desired manipulation, \( \mathbf{v_n} \) is a negative attribute value, \( d \) is the distance metric, and \( \beta \) is a margin that separates the positive and negative attribute embeddings. This loss function ensures that the modified attributes are accurately represented according to the target label, promoting effective feature alignment.
\end{itemize}

We can summarize the overall architecture as follows. The approach starts with the disentangled feature extractor of ADDE \cite{hou2021ADDE}, which leverages a CNN-based backbone followed by per-attribute encoders.
The attribute representations are then processed by a transformer encoder, followed by a transformer decoder that includes information coming from a parallel memory branch. In the memory branch, prototype embeddings for various attribute values are processed with an additional transformer encoder, which applies targeted changes to the identified attribute based on the manipulation vector.
To guide the model's training, we utilized two loss functions, namely the Compositional Triplet Loss and Label Triplet Loss. These losses aim to refine the attribute manipulation process by ensuring accurate feature alignment and consistency with target attributes.

Once the target feature is obtained, we perform knn retrieval in a gallery set, retrieving garments with FAISS~\cite{douze2024faiss}.

\section{Experimental Protocol}
\label{sec:exp}

\subsection{Datasets}
Experiments were conducted with two datasets commonly used for the task of fashion attribute manipulation:
\textbf{Shopping100k} \cite{shopping100k} and \textbf{DeepFashion} \cite{deepfashion}. Shopping100k comprises 101,021 images of clothing, each with dimensions of 762 $\times$ 1100 pixels. Each image is annotated with at least 5 out of 12 possible attributes: collar, color, fabric, fastening, fit, gender, length, neckline, pattern, pocket, sleeve length, and sport. These attributes combine to form 151 unique labels, providing detailed descriptions. Sourced from various e-commerce platforms, the dataset includes 15 high-level apparel categories, offering a diverse and comprehensive set of fashion items for analysis. 

DeepFashion includes a vast collection of over 800,000 fashion images, ranging from well-posed shop samples to informal consumer pictures. It supports four key benchmarks: Attribute Prediction, Consumer-to-Shop Clothes Retrieval, In-Shop Clothes Retrieval, and Landmark Detection. Each image is extensively annotated with 50 categories, 1,000 descriptive attributes, as well as bounding boxes and detailed clothing landmarks. For Shopping100k, 80,586 images were utilized for training and 20,000 images for testing. Regarding DeepFashion, 3 of the 6 available attributes (category, texture, and shape) were selected, in line with prior works \cite{ak2018FSN,hou2021ADDE}. Under-represented attribute values were excluded, as well as images that possessed multiple attribute labels for each type. For both datasets, 2,000 query images were employed for the attribute manipulation task. Every possible manipulation was applied to each query image.

\subsection{Evaluation Metrics}
\label{sec:metrics}
To assess the effectiveness of the proposed framework, two evaluation metrics are employed: top-K retrieval accuracy (with $K$ values of 10, 20, and 30) and Normalized Discounted Cumulative Gain (NDCG@K) \cite{jarvelin2002cumulated} with $K$ set to 30. Top-K retrieval accuracy is defined as the ratio of "hit" queries to the total number of queries. A query is considered a "hit" if at least one image among the top-K nearest neighbors possesses the expected attributes. This metric is mathematically expressed as follows:

\begin{equation}
\text{top-K Retrieval Accuracy} = \frac{\text{Number of hit queries}}{\text{Total number of queries}}
\end{equation}

\noindent The NDCG@K metric is used to evaluate ranking quality and is defined as:

\begin{equation}
    \text{NDCG@K} = \frac{1}{Z} \sum_{j=1}^{K} \frac{2^{\text{rel}(j)} - 1}{\log_2(j + 1)}
\end{equation}

In this formula, $rel(j)$ represents the relevance score of the $j$-th retrieved item. This score is computed by determining the number of matching attributes between the desired label and the actual label of the $j$-th ranked item, divided by the total number of attribute types. The normalization factor $Z$ ensures the appropriate scaling of the NDCG@K value. To further evaluate the system's ability to preserve unchanged attributes, two variants of the NDCG metric are introduced: NDCG\textsubscript{target} and NDCG\textsubscript{others}. Although these variants follow the general formula of the standard NDCG, they differ in their methods for calculating relevance scores $rel(j)$. NDCG\textsubscript{target} specifically measures the target attribute intended for modification, while NDCG\textsubscript{others} evaluates the attributes that are meant to remain unchanged.

\section{Results and Discussion}
\label{sec:res}
\subsection{Comparative Performance Analysis}
In the following we report a quantitative analysis of the performance of \short{} on both the Shopping100k and the DeepFashion datasets.

\paragraph{\textbf{Results on Shopping100k}}
Table \ref{tab:comp_shopping100k_topk} reports the top-K retrieval accuracies of the proposed model on the Shopping100k dataset and compares these results with state-of-the-art methods. As the number of retrieved items increases, accuracy improves, reaching over 60.79 at $K$=30. The results are benchmarked against several baselines, including AMNet \cite{zhao2017AMNet}, FSN \cite{ak2018FSN}, ADDE-M \cite{hou2021ADDE} and ADGR \cite{scaramuzzino2023attribute}. Notably, most methods, including the proposed one, employ an AlexNet \cite{alexnet} backbone for feature extraction, thereby ensuring a fair comparison. The only exception is DAtRNet \cite{bhattacharya2022datrnet}, which employs a custom backbone. We include this method in the results albeit not directly comparable with the other approaches, as advantaged by the feature extractor.
As shown in Table \ref{tab:comp_shopping100k_topk} our proposed approach (\short{}) outperforms the previous methods over all the top-K retrieval metrics. In particular, the biggest gain compared to ADDE, on which we built our system, can be observed at Top-10, i.e. the most challenging setting.

\begin{table}[t]
\centering
\caption{State-of-the-art comparisons on Shopping100k in terms of top-K accuracy.}
\label{tab:comp_shopping100k_topk}
\begin{tabular}{lccccHH}
\toprule
\textbf{Method} & ~~Backbone~~ & ~~\textbf{Top-10}~ & ~\textbf{Top-20}~ & ~\textbf{Top-30} & \textbf{Top-40} & \textbf{Top-50}\\
\midrule
DAtRNet \cite{bhattacharya2022datrnet} & Custom & - & - & 67.70 & - & -\\
\midrule
AMNet \cite{zhao2017AMNet} & AlexNet & 25.62 & 36.13 & 42.94 & 47.71 & 51.64\\
FSN \cite{ak2018FSN} & AlexNet & 38.41 & 47.44 & 57.17 & 61.62 & 66.70\\
ADDE-M \cite{hou2021ADDE} & AlexNet & 41.17 & 52.93 & 59.81 & 64.10 & 67.29\\
ADGR \cite{scaramuzzino2023attribute} & AlexNet & 43.02 & 54.13 & 60.76 & \textbf{65.18} & 68.53\\
\textbf{\short{}} & AlexNet & \textbf{43.70} & \textbf{54.37} & \textbf{60.79} & 64.94 & \textbf{68.64}\\
\bottomrule
\end{tabular}
\end{table}

\begin{table}[t]
\centering
\caption{State-of-the-art comparisons on Shopping100k in terms of NDCG@30.}
\label{tab:comp_shopping100k_ndcg}
\begin{tabular}{lccc}
\hline
\textbf{Method} & ~~~\textbf{NDCG@30}~ & ~\textbf{NDCG\textsubscript{t}@30}~ & ~\textbf{NDCG\textsubscript{o}@30}~ \\
\hline
ADDE-M \cite{hou2021ADDE} & 73.67 & 43.05 & 77.79 \\
\textbf{\short{}} & \textbf{73.88} & \textbf{43.75} & \textbf{77.91}\\
\hline
\end{tabular}
\end{table}

Beyond the top-K retrieval accuracy, Table \ref{tab:comp_shopping100k_ndcg} provides a comparison in terms of NDCG@30 for the Shopping100k dataset against ADDE-M \cite{hou2021ADDE}.
The proposed approach outperforms ADDE-M \cite{hou2021ADDE} across the three NDCG metrics, with the most significant improvement observed in NDCG\textsubscript{target}, reaching 43.75 against the 43.05 baseline. This hints at the fact that \short{} is capable of better manipulating garments to include the desired target attribute, while being able to better retain all other attributes unaltered.

\paragraph{\textbf{Results on DeepFashion}}
Tables \ref{tab:comp_deepfashion_topk} and \ref{tab:comp_deepfashion_ndcg} show, respectively, the top-K retrieval accuracy and NDCG@30 metrics for the DeepFashion database. As evidenced by the results in Table \ref{tab:comp_deepfashion_topk}, the proposed method the proposed method still outperforms the other methods over the Top-10 and Top-20 metrics, losing by only 0.08 points to  \cite{hou2021ADDE} in the Top-30 category.
It is worth noticing that in general the DeepFashion dataset is more challenging than Shopping100k. In fact, the images depict persons wearing the fashion items, thus making it harder for vision models to interpret each attribute due to clutter and possible occlusions. This is not the case in Shopping100k, where images are typically shown on a uniform background.

\begin{table}[t]
\centering
\caption{State-of-the-art comparisons on DeepFashion in terms of top-K accuracy.}
\label{tab:comp_deepfashion_topk}
\begin{tabular}{lcccHH}
\toprule
\textbf{Method} & ~~\textbf{Top-10}~~ & ~~\textbf{Top-20}~~ & ~~\textbf{Top-30} & ~~\textbf{Top-40}~~ & ~~\textbf{Top-50}~~\\
\midrule
AMNet \cite{zhao2017AMNet} & 14.11 & 19.39 & 22.94 & 25.51 & 27.58\\
ADDE-M \cite{hou2021ADDE} & 23.60 & 28.58 & \textbf{31.52} & \textbf{33.98} & \textbf{35.91}\\
\textbf{\short{}} & \textbf{23.69} & \textbf{28.73} & 31.44 & 33.74 & 35.67\\
\bottomrule
\end{tabular}
\end{table}

\begin{table}[t]
\centering
\caption{State-of-the-art comparisons on DeepFashion in terms of NDCG@30.}
\label{tab:comp_deepfashion_ndcg}
\begin{tabular}{lccc}
\hline
\textbf{Method} & ~~~~\textbf{NDCG@30}~~ & ~~\textbf{NDCG\textsubscript{t}@30}~~ & ~~\textbf{NDCG\textsubscript{o}@30} \\
\hline
ADDE-M \cite{hou2021ADDE} & 32.91 & \textbf{34.70} & 36.29\\
\textbf{\short{}} & \textbf{33.75} & 33.07 & \textbf{38.58}\\
\hline
\end{tabular}
\end{table}

\begin{table}[t]
\centering
\caption{Results for the ablation studies in terms of top-K over Shopping100k.}
\label{tab:ablation}
\begin{tabular}{lcccHH}
\toprule
\textbf{Method} & ~~\textbf{Top-10}~~ & ~~\textbf{Top-20}~~ & ~~\textbf{Top-30} & ~~\textbf{Top-40}~~ & ~~\textbf{Top-50}~~\\
\midrule
\textbf{GAMMA} & \textbf{43.70} & \textbf{54.37} & \textbf{60.79} & 64.94 & \textbf{68.64}\\
Encoder-Decoder & 43.04 & 54.03 & 60.55 & & \\ 
Encoder & 38.80 & 50.75 & 57.43 & & \\ 
\bottomrule
\end{tabular}
\end{table}
\subsection{Ablation Studies}
To assess the effectiveness of each component within our proposed architecture, we conducted a series of ablation studies. These studies focused on evaluating how different configurations, particularly the inclusion of the memory block and the overall encoder-decoder design, influence the model's performance in fashion image retrieval tasks.
We evaluated three primary model configurations:
\begin{itemize}
    \item \textbf{GAMMA} (Dual-Encoder + Memory Block): Our baseline model employs separate encoders for the two inputs: the query image embedding and the target embedding from the memory block. These are followed by a final decoder, as detailed in Section \ref{sec:method}.
    \item \textbf{Encoder-Decoder Model}:  This variant features a single encoder that processes the query image features along with the manipulation vector, formulated as: \begin{equation}
    \mathbf{r_F} = \text{Encoder}(PE([\mathbf{r_1,r_2...,r_n}, MLP(\mathbf{i})]))
    \end{equation}

    where \( \mathbf{r_1,r_2...,r_n} \) denote the disentangled representation of the query image and \( \mathbf{i} \) the manipulation vector. The decoder then uses the encoded features and the target embedding from the memory block to generate the final output \( \mathbf{r'} \). 
    \begin{equation} \mathbf{r'} = \text{Decoder}(\mathbf{r_F},\mathbf{M_{\mathbf{p}}})
    \end{equation}
    where \(\mathbf{M_{\mathbf{p}}}\) represents the prototype target embedding from the memory block.
    \item \textbf{Single Encoder Model}: A simplified architecture where the backbone features along with the manipulation vector are processed by a single encoder. This model does not use a decoder or memory block embeddings and can be formulated as: 
    \begin{equation}
    \mathbf{r'} = \text{Encoder}(PE([\mathbf{r_1,r_2...,r_n}, MLP(\mathbf{i})]))
    \end{equation}

\end{itemize}

As reported in Table \ref{tab:ablation}, the proposed approach (GAMMA) achieved the highest performance across all top-K retrieval metrics. The Encoder-Decoder model, while performing better than the Single Encoder variant, shows a performance reduction, particularly at higher K values, indicating that the separation of attribute and content features is crucial for retaining attribute specificity.
Overall, these studies confirm that both the dual-encoder architecture and the memory block are essential for the superior performance of our approach, enabling effective and precise manipulation of fashion image attributes.

\subsection{Qualitative Results} 
\begin{figure}[t]
    \centering
\includegraphics[width=\linewidth]{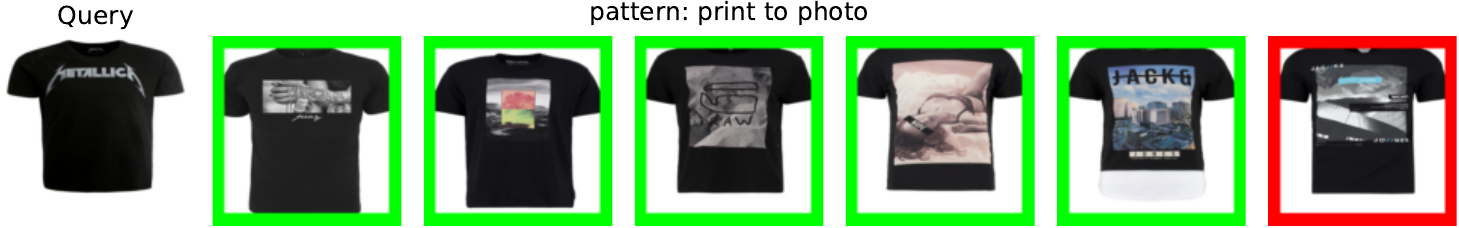}\\~\\
\includegraphics[width=\linewidth]{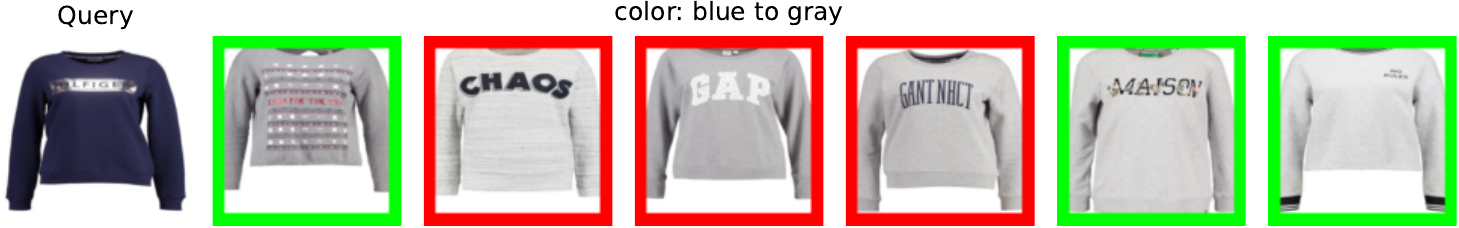}\\~\\
\includegraphics[width=\linewidth]{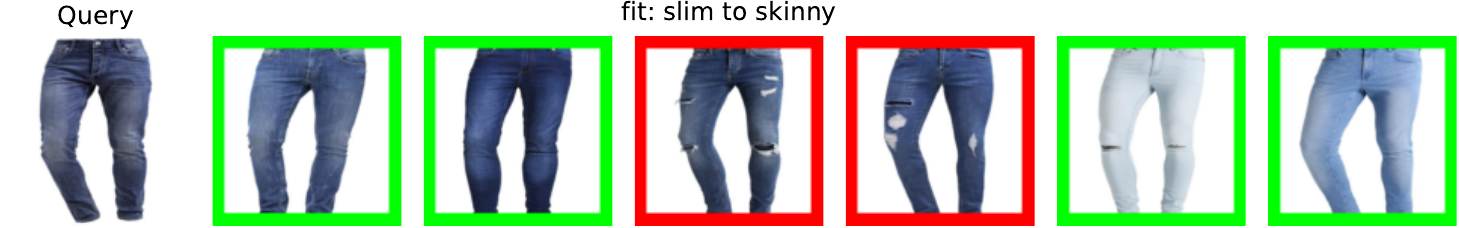}\\~\\
\includegraphics[width=\linewidth]{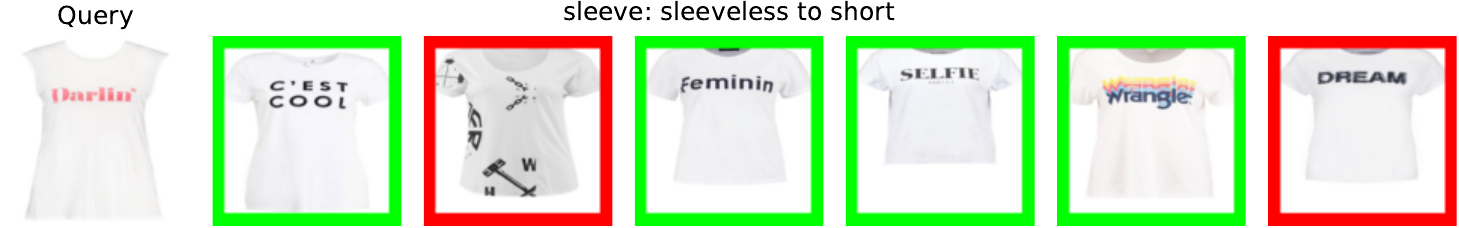}\\~\\
\includegraphics[width=\linewidth]{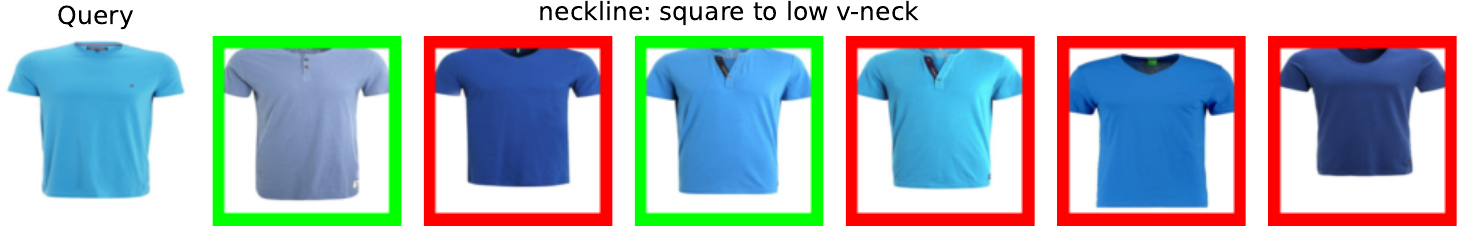}\\~\\

    \caption{Qualitative results over the Shopping100k testset. Left most column shows the query image. The rest are sorted retrieved results of our model. Green outline represent the ground truth correct answers.}
    \label{fig:qualitative}
\end{figure}

In this subsection we show qualitative results along with some failure cases. In Figure \ref{fig:qualitative} we report some sample results for the retrieval task conditioned on both a query image and a modified attribute. The first column of Figure \ref{fig:qualitative} shows the query image and the captions describe the manipulation to apply to the input fashion item.
The retrieved results are color-coded according to their ground truth value in the Shopping100k test set, with a green border for correctly retrieved results and a red border for results that do not match the desired attributes. Looking at these samples it appears that the model is capable of manipulating disentangled attributes for each item, i.e. retrieving the best items such that one and only one attribute is changed. It is worth noticing that even the failure cases are graceful failures, i.e. they maintain most of the desired attributes and look visually similar.

\begin{figure}[t]
    \centering
\includegraphics[width=\linewidth]{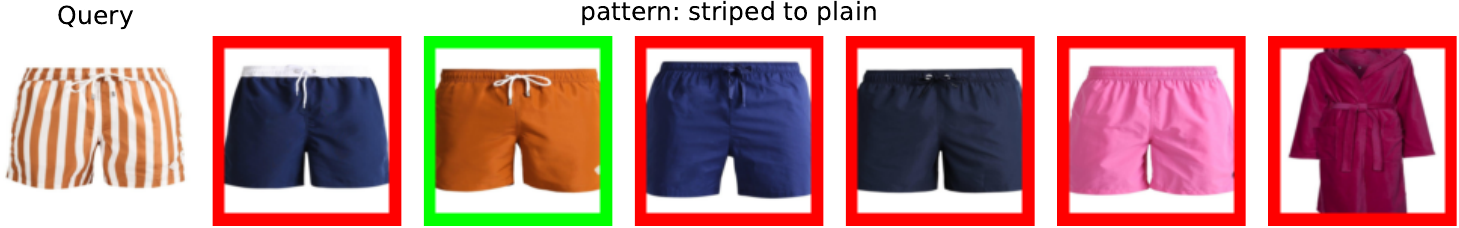}\\~\\
\includegraphics[width=\linewidth]{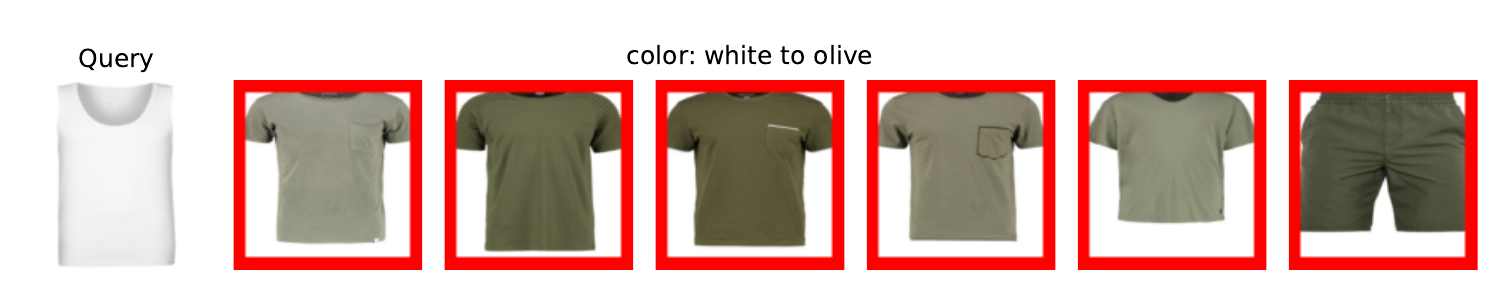}\\~\\
\includegraphics[width=\linewidth]{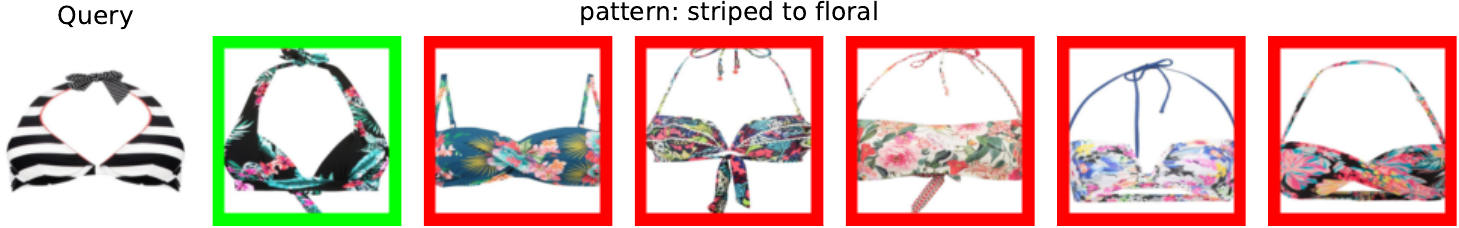}\\~\\

    \caption{Failure cases over the Shopping100k testset. Left column shows the query image, the others are ranked results. Green outline represents the ground truth correct answers.}
    \label{fig:fail}
\end{figure}

In Figure \ref{fig:fail} instead we show some failure cases for our model. In the case of the first row, the model correctly retrieves similar garments and changes the style from \textit{striped} to \textit{plain} but does not keep the color consistent. The second row shows a similar case in which the color is correctly changed but the sleeve length is mistakenly altered. Finally, in the third sample, it appears that our model correctly retrieved the top six items but the original annotations do not agree.

\section{Conclusions}
\label{sec:conc}
\vspace{-8pt}
In the task of generating attribute-specific modifications in clothing images, the proposed architecture offers a systematic approach to feature extraction and manipulation useful for item retrieval. Our proposed model (GAMMA) integrates several components to build a solution capable of receiving a query image and an attribute manipulation to produce a manipulated representation that is then used to retrieved the desired item.
By employing a dual-encoder architecture combined with a memory block, our approach adeptly handles the disentanglement and targeted modification of image attributes. 
The experimental results demonstrate that GAMMA outperforms existing state-of-the-art methods in top-K retrieval accuracy and NDCG metrics on both the Shopping100k and DeepFashion datasets. The ablation studies confirm the importance of our dual-encoder and memory block design, highlighting their role in achieving superior performance.
Looking ahead, future work will focus on extending GAMMA's capabilities to support query language-based attribute manipulation. This involves integrating natural language processing techniques to enable users to specify attribute changes through textual queries, further enhancing the model's usability and flexibility.

\section{Acknowledgements}
This work was partially supported by the European Commission under European Horizon 2020 Programme, grant number 951911—AI4Media
\bibliographystyle{splncs04}
\bibliography{main}
\end{document}